\documentclass{article}

    \PassOptionsToPackage{numbers, sort&compress}{natbib}

\usepackage[preprint]{preprint}

\usepackage[utf8]{inputenc} %
\usepackage[T1]{fontenc}    %
\usepackage[colorlinks=true,linkcolor=teal,citecolor=olive]{hyperref} %
\usepackage{url}            %
\usepackage{booktabs}       %
\usepackage{amsfonts}       %
\usepackage{nicefrac}       %
\usepackage{microtype}      %
\usepackage{xcolor}         %
\usepackage{lipsum}
\usepackage{amsmath}
\usepackage{amssymb}
\usepackage{mathtools}
\usepackage{amsthm}
\usepackage{svg}
\usepackage{multirow}
\usepackage{siunitx}
\usepackage{placeins}
\usepackage{caption}
\usepackage[capitalise,nameinlink]{cleveref}
\usepackage{subcaption}
\usepackage{tabularx}
\usepackage{wrapfig}
\usepackage{graphicx}
\usepackage{makecell}
\usepackage[misc]{ifsym}
\usepackage{bbm}
\usepackage{enumitem}
\usepackage{pifont}
\usepackage{gensymb}
\fontsize{10}{12}\selectfont %

\title{Wearable data from subjects playing Super Mario, sitting university exams, or performing physical exercise help detect acute mood episodes via self-supervised learning
}

\newcommand\blfootnote[1]{%
    \begingroup
    \renewcommand\thefootnote{}\footnote{#1}%
    \addtocounter{footnote}{-1}%
    \endgroup
}

\author{
    Filippo Corponi$^{1}$ \quad Bryan M. Li$^{1}$  \quad Gerard Anmella$^{2}$ 
    \quad Clàudia Valenzuela-Pascual$^{2}$\\[0.5mm]
    \textbf{\quad Ariadna Mas$^{2}$
    \quad Isabella Pacchiarotti$^{2}$ \quad Marc Valentí$^{2}$ \quad
    Iria Grande$^{2}$} \\[0.5mm]
    \textbf{\quad Antonio Benabarre$^{2}$ \quad Marina Garriga $^{2}$
    \quad Eduard Vieta$^{2}$ \quad Allan H Young$^{3}$
    } \\[0.5mm]
    \textbf{
    \quad Stephen M. Lawrie$^{4}$ \quad Heather C. Whalley$^{4}$ \quad Diego Hidalgo-Mazzei$^{2}$ \quad Antonio Vergari$^{1}$} \\[2mm]
    $^1$School of Informatics, University of Edinburgh\\
    $^2$Hospital Cl\'{i}nic de Barcelona, University of Barcelona\\
    $^3$Institute of Psychiatry, Psychology and Neuroscience, King’s College London\\
    $^4$Division of Psychiatry, University of Edinburgh
}

\begin{document}

\maketitle

\begin{abstract}

    Personal sensing, leveraging data passively and near-continuously collected with wearables from patients in their ecological environment, is a promising paradigm to monitor mood disorders (MDs), a major determinant of worldwide disease burden. However, collecting and annotating wearable data is very resource-intensive. Studies of this kind can thus typically afford to recruit only a couple dozens of patients. This constitutes one of the major obstacles to applying modern supervised machine learning techniques to MDs detection.
    
    In this paper, we overcome this data bottleneck and advance the detection of MDs acute episode vs stable state from wearables data on the back of recent advances in self-supervised learning (SSL). This leverages unlabelled data to learn representations during pre-training, subsequently exploited for a supervised task.
    
    First, we collected open-access datasets recording with an Empatica E4 spanning different, \textit{unrelated to MD monitoring}, personal sensing tasks -- from emotion recognition in Super Mario players to stress detection in undergraduates -- and devised a pre-processing pipeline performing on-/off-body detection, sleep-wake detection, segmentation, and (optionally) feature extraction. With 161 E4-recorded subjects, we introduce E4SelfLearning, the largest to date open access collection, and its pre-processing pipeline~\footnote{\href{https://github.com/april-tools}{repository to be released upon acceptance for publication}}. Second, we show that SSL confidently outperforms fully-supervised pipelines using either our novel E4-tailored Transformer architecture (E4mer) or classical baseline XGBoost: 81.23\% against 75.35\% (E4mer) and 72.02\% (XGBoost) correctly classified recording segments from 64 (half acute, half stable) patients. Lastly, we illustrate that SSL performance is strongly associated with the specific surrogate task employed for pre-training as well as with unlabelled data availability.

    \blfootnote{\Letter\ Correspondence to: \href{mailto:filippo.corponi@.ed.ac.uk}{\texttt{filippo.corponi@.ed.ac.uk}} 
    }
        
    \noindent\textbf{Keywords:} mood disorders; time-series classification; wearables; personal sensing; deep learning; self-supervised learning; transformer
    
\end{abstract}

\section{Introduction} \label{introduction}

Mood disorders (MDs) are a group of mental health conditions in the Diagnostic and Statistical Manual $5\textsuperscript{th}$ edition (DSM-5)~\cite{american2013diagnostic} classification system. They are chronic, recurrent disorders featuring disturbances in emotions, energy, and thought, standing out as a leading cause of worldwide disability \citep{santomauro2021global, greenberg2021economic} and suicidality \citep{braadvik2018suicide}. Timely recognition of mood episodes is critical towards better outcomes \cite{joyce2016treatment}. However, this is challenging due to generally limited patient insight \cite{buchman2020paradoxical} compounded with low availability of specialized care for MDs, with rising demand straining current capacity \cite{rimmer2021mental, satiani2018projected}.

Personal sensing, involving the use of machine learning (ML) to harness data passively and near-continuously collected with wearable devices from patients in their ecological environment, has been attracting interest as a promising paradigm to address this gap \citep{mohr2017personal}. 
Indeed, some of the core MD clinical features (e.g. disturbance in mood, energy levels) translate into changes in physiological parameters measurable with wearable devices \cite{faurholt2017state, sarchiapone2018association, tazawa2019actigraphy}. 
A major barrier towards the development of clinical decision support systems featuring personal sensing has been the scarcity of labelled data, that is data with annotations by clinicians about the MD state (e.g. diagnosis, disease phase, symptoms severity). 
Collecting and annotating data for personal sensing in MDs is in fact an expensive and time-consuming enterprise; thus, studies typically use samples running into only few dozens of patients \cite{jacobson2019digital,tazawa2019actigraphy,cote2022long, nguyen2022decision, ghandeharioun2017objective, pedrelli2020monitoring, lee2022prediction, corponi2023automated}. 

\begin{figure}[!t]
    \centering
    \includegraphics[width=1\linewidth]{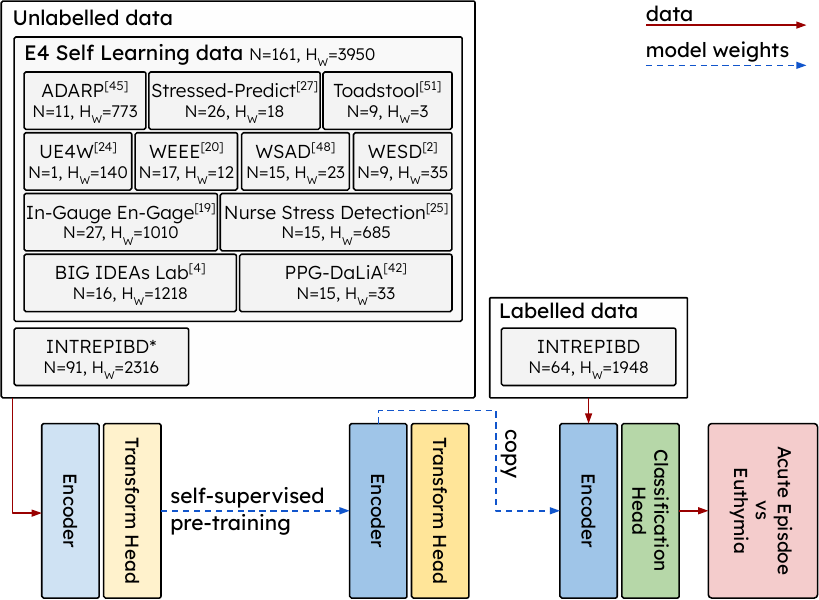}
    \caption{\textbf{A total of 6267 hours ($\sim$261 days) of unlabelled recordings from 252 subjects while awake were used for self-supervised pre-training}. Unlabelled data comprised a collection of eleven open-access datasets, whose pre-processed version we make publicly available (E4SelfLearning), along with part of the INTREPIBD study that was not relevant for the target task under investigation, i.e. acute episode vs euthymia classification. Unlabelled data was passed through a model consisting of an encoder and a transform head for self-supervised pre-training; the pre-trained encoder block was then retained for the target task while the transform head was replaced with a new, randomly initialized classification head. N: subjects \#; H$_{\text{w}}$: waking hours \#; $^*$ figures herewith reported for the INTREPIBD study do not include the target task (labelled) training set which was also used during self-supervised pre-training.
    }
    \label{figure:datasets}
\end{figure}

In this work, we take a different perspective and leverage \textit{unlabelled} data collected with the Empatica E4 wristband~\citep{e4spec}, a popular research-grade device for personal sensing studies \cite{ronca2023wearable}, as well as recent advancements in self-supervised learning (SSL) techniques that can learn meaningful representations from such unlabelled data. 
Specifically, we take advantage of a number of open-access datasets which record physiological data with the E4 across different settings but do not address MDs and therefore do not provide information about the mood state of the subjects involved. 
While each such dataset has only a limited number of subjects, our aggregated and preprocessed dataset E4SelfLearning can break the labelled data bottleneck for personal sensing in MDs (\Cref{figure:datasets}).

Fully-supervised systems require vast amounts of data to train, thus limiting their application in fields, such as healthcare, where amassing large, high-quality datasets is demanding in terms of time and human resources \cite{shani2023lean}. While previous studies in personal sensing for MDs investigated different tasks, including acute episode detection \cite{jacobson2019digital,tazawa2019actigraphy,cote2022long, nguyen2022decision}, regression of a psychometric scale total score \cite{ghandeharioun2017objective, pedrelli2020monitoring, lee2022prediction}, and more recently multi-task inference of all items in two commonly used psychometric scales \cite{li2022inferring, corponi2023automated}, they all developed their models in a fully-supervised fashion, i.e. they were trained on samples for which ground-truth labels were available. As a result, considering that obtaining clinical annotations from patients, especially when on an acute MD episode, is a challenging and expensive enterprise, the sample size is generally modest (e.g. 52 in \citealt{cote2022long}, 45 in \citealt{tazawa2019actigraphy}, or 31 \citealt{pedrelli2020monitoring}).

SSL on the other hand is a framework where the model learns from unlabelled data without any need for external supervision, therefore alleviating the annotation bottleneck and allowing us to repurpose existing unlabelled datasets \cite{rani2023self}.
Indeed, SSL derives supervisory signal from the data itself thanks to pretext tasks, that is new supervised challenges, for example imputing occluded parts of the input data. 
Through such preparatory pretext tasks, not requiring expert annotation, the model learns useful representations, partial solutions to the down-stream target task of interest, for which only a comparatively small amount of annotated data is available \citep{ericsson2022self}. On the back of the great success of SSL in Computer Vision (CV) \cite{ohri2021review} and Natural Language Processing (NLP) \cite{devlin2018bert}, and with encouraging findings in other healthcare applications \cite{huang2023self}, we extend pioneering SSL works on multi-variate time-series \cite{zhang2023self,zerveas2021transformer, wu2023transformer} to personal sensing in MDs.

In this work, we make the following contributions:

\begin{itemize}
    \item We gather eleven open-access datasets recording physiological data with an Empatica E4 device and developed a pipeline for pre-processing such data that does on-/off-body detection, sleep-wake detection, segmentation, and (optionally) feature-extraction. We make the pre-processing pipeline and the pre-processed data publicly available. This collection (E4SelfLearning), with 161 subjects, is the biggest open-access to date. We believe that this effort can stimulate future research into SSL with multi-variate time-series sensory data by removing two barriers, pre-processing and data availability.
    \item 
    We propose a novel Transformer \cite{vaswani2017attention} architecture (E4mer, \Cref{figure:E4mer}) and show that SSL is a viable paradigm, outperforming both the fully-supervised E4mer and a state-of-the-art classical machine learning (CML) model using handcrafted features in distinguishing MD acute episode from clinical stability (euthymia in psychiatric parlance), i.e. a time-series (binary) classification task.
    \item We investigate what makes SSL successful. Specifically, we compare two main pretext task designs (i.e. masked prediction and transformation prediction) \cite{ericsson2022self} and for the best performing routine we study its sensitivity to the unlabelled data availability in ablation analyses. We inspect learned embeddings and show that they capture meaningful semantics about the underlying context, i.e. sleep-wake status and symptoms severity.
    
\end{itemize}

\section*{Results}

\subsection*{Study sample \& data preparation}

Our target task is distinguishing MD acute episode from euthymia using wearable data. We start from a dataset for which we have labelled samples, the INTREPIBD cohort \cite{anmella2023exploring}: 64 individuals with a DSM-5 MD diagnosis (either Major Depressive Disorder (MDD) or Bipolar Disorder (BD)), half euthymic (score $\leq7$ on the Hamilton Depression Rating Scale-17 \cite{hamilton1960rating} and Young Mania Rating Scale \cite{young1978rating} for at least 8 weeks \cite{tohen2009international}) and half at the onset of an acute episode (of manic, depressive, or mixed polarity) according to DSM-5 criteria. Some patients from the INTREPIBD study recruited at the onset of an acute mood episode have up to three follow-up assessments and recordings; follow-ups excluded from all analysis herewith presented.

Patients were interviewed by a psychiatrist collecting clinical-demographics (\Cref{table:target_task_cohort}) and were required to wear on their non-dominant wrist an E4 device until battery ran out ($\sim$48 hours). 

\begin{figure}[!t]
    \centering
    \includegraphics[width=1\linewidth]{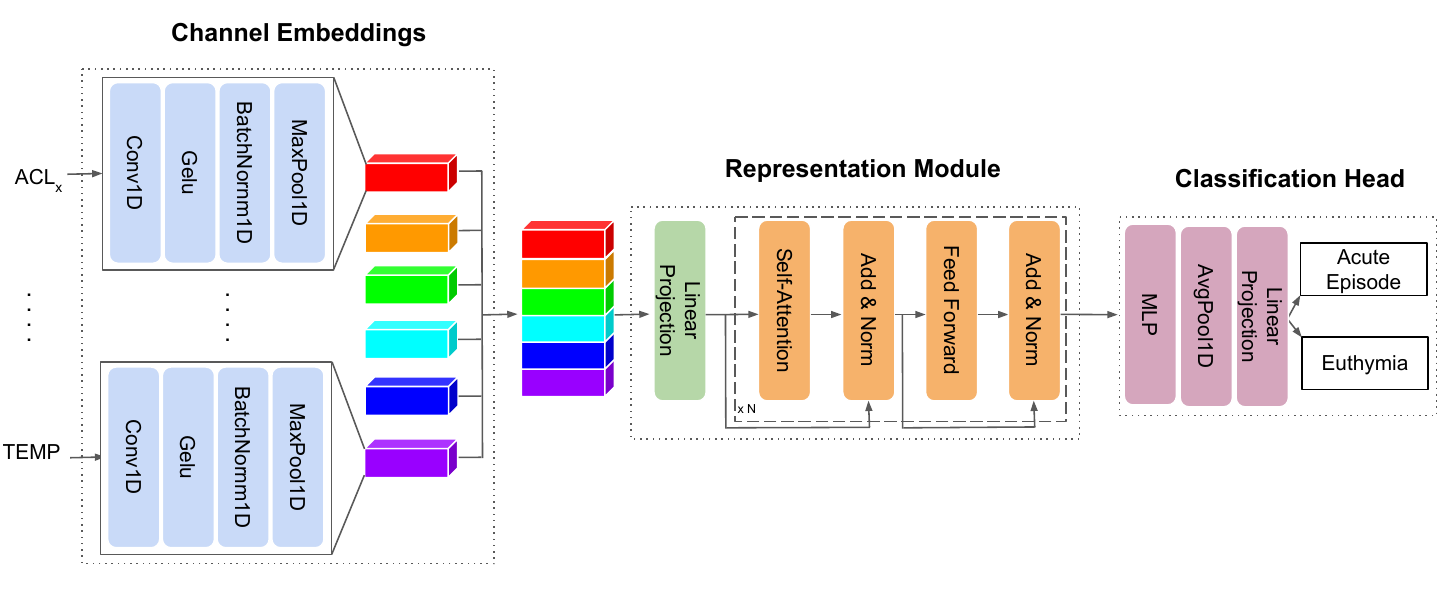}
    \caption{\textbf{E4mer is a Transformer model tailored to the Empatica E4 input data}. The E4mer is constituted of three sequential modules: 1) Channel embeddings set in parallel, one for each Empatica E4 raw input channel (i.e. ACL$_{\mathsf{x}}$, ACL$_{\mathsf{y}}$, ACL$_{\mathsf{z}}$, BVP, EDA, TEMP), extracting features and mapping channels to tensors of dimensionality ($B$=batch size, $N$= time steps, $F$= filters \#) so that they can be conveniently concatenated along the dimension $F$; 2) Representation Module learning contextual representations of the input time steps within the input segment thanks to the multi-head self-attention mechanism; 3) Classification Head outputting probabilities for the two target classes, i.e. acute episode and euthymia. Self-supervised learning models employed in our experiments feature the same E4mer architecture described above, where the Classification Head however is replaced with a Transform Head projecting onto a label space compatible with the pretext task at hand.
    }
    \label{figure:E4mer}
\end{figure}

This wearable records (sampling rate) 3D acceleration (ACL, 32Hz), blood volume pressure (BVP, 64Hz), electrodermal activity (EDA, 4Hz), heart rate (HR, 1Hz), inter-beat intervals (IBI, i.e. the time between two consecutive heart ventricular contractions) and skin temperature (TEMP, 1Hz). Pre-processing steps were, sequentially: on-/off-body detection, sleep/wake detection, and segmentation. Note that segments coming from the same recording all shared the same ground truth label with respect to MD state. Off-body and sleep segments were discarded from supervised as well as self-supervised training. As HR and IBI are not raw features but derivations of BVP through a proprietary algorithm, these were excluded from segments inputted to the E4mer (\Cref{figure:E4mer}) models. As for CML modelling, handcrafted features were extracted from segments with the popular open-access toolkit \textit{FLIRT} \cite{foll2021flirt}. An equal number of segments for each class (acute episode and euthymia) was retained and segments were split into train/val/test with a proportion of 70:15:15 along recording time.

For self-supervised pre-training, we gathered eleven open-access datasets recording with an E4 \cite{sah2022adarp, reiss2019deep_ppg_dalia, iqbal2022stress_sps, svoren2020toadstool, gashi2022multidevice_weee, schmidt2018introducing_wesad, amin2022wearable_wesd, hosseini2022multimodal_hospital_nurses, gao2022understanding_in_gauge, UE4W}. 
While they all used the same wearable device, such datasets could differ substantially with respect to population, recording setting, and task: from students taking exams \cite{amin2022wearable_wesd} or attending classes \cite{gao2022understanding_in_gauge}, to nurses carrying out their duty \cite{hosseini2022multimodal_hospital_nurses} and subjects performing different physical activities \cite{gashi2022multidevice_weee} or playing Super Mario \cite{svoren2020toadstool}. Subjects that are not part of the target classes from the INTREPIBD study are also included in the unlabelled data for SSL. Unlabelled recording sessions underwent the same pre-processing scheme explained above, producing a total of 6267 waking hours from 252 individuals, and were split into train/val with a ratio of 85:15. While pre-training does not require assessing generalization performance on a test set and thus unlabelled data was only split into train/val, the target task validation set was recycled as a test set for evaluating generalization performance on the pretext tasks.

\begin{table}[!t]
    \caption{\textbf{Clinical-demographic features of target task (acute episode vs euthymia classification) population.} Mood episodes clinically lie on a spectrum, with depression on one end and mania on the other; mixed episodes, featuring symptoms from both polarities, are a bridge between the two spectrum extremes. In this study we considered acute episode of any polarity and similarly we considered euthymia as a unique class, whether in the context of a bipolar or major depression diagnosis. BD: Bipolar Disorder; HDRS: Hamilton Depression Rating Scale-17; MDD (Major Depressive Disorder); MDE-BD (Major Depressive Episode in Bipolar Disorder); MDE-MDD (Major Depressive Episode in Major Depressive Disorder); ME (Manic Episode); MX (Mixed Episode); YMRS: Young Mania Rating Scale.
    } \label{table:target_task_cohort}
    \begin{center}
    \begin{small}
    \begin{sc}
    \setlength{\tabcolsep}{3.5pt}
    \begin{tabular}{lcccccc}
        \toprule
        & Age & Females & Diagnosis & HDRS & YMRS\\
        & mean (std)  & N (percentage) & & mean (std) & mean (std) \\
        \textbf{Euthymia}& 47.22 (16.06) & 14 (43.75\%) & BD (N=26) & 2.93 (1.73) & 1.3 (1.61) \\
        N=32 &  & & MDD (N=6) & 3.14 (1.95)  & 0.29 (0.76) \\
        \textbf{Acute episode} & 50.56 (13.05) & 15 (46.88\%) & MDE-BD (N=9) & 20.22 (6.34) & 2.56 (3.94) \\   
        N=32  & & & MDE-MDD (N=7) & 25.14 (4.78) & 1.86 (2.41) \\  
        & & & ME (N=14) & 5.67 (4.37) & 20.13 (6.28) \\  
        & & & MX (N=2) & 16 (4.24) & 13.5 (4.95) \\

        \bottomrule
    \end{tabular}
    \end{sc}
    \end{small}
    \end{center}
\end{table}
\noindent

\subsection*{Surrogate tasks used in self-supervised pre-training}

The same model, using the E4mer architecture, was used across different pretext tasks, that is an encoder $\mathsf{EN}$, consisting of convolutional channel embeddings $\mathsf{CEs}$, whose output had constant dimensionality regardless of the input channel sampling frequency to allow embeddings concatenation, and a Transformer \cite{vaswani2017attention} representation module $\mathsf{RM}$, followed by a simple multi layer perceptron (MLP) functioning as transform head $\mathsf{H_{ssl}}$. 

\begin{figure}[!t]
    \centering
    \begin{subfigure}[b]{0.85\textwidth}
       \includegraphics[width=1\linewidth]{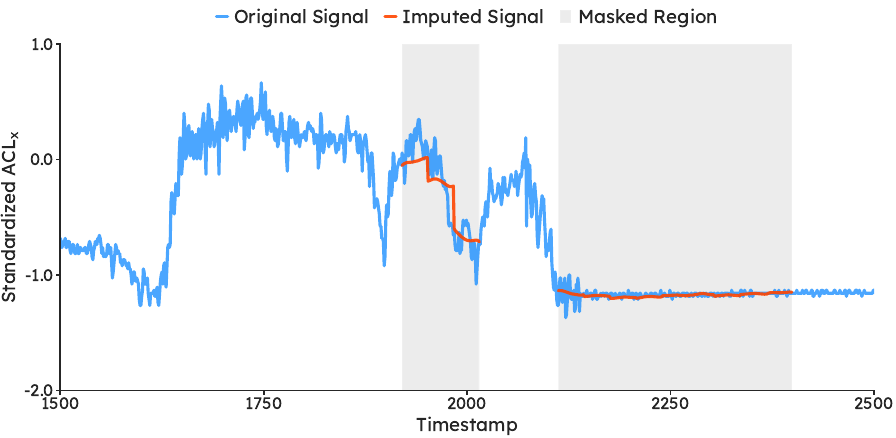}
       \caption{}
       \label{fig:denoising} 
    \end{subfigure}
    \begin{subfigure}[b]{0.85\textwidth}
       \includegraphics[width=1\linewidth]{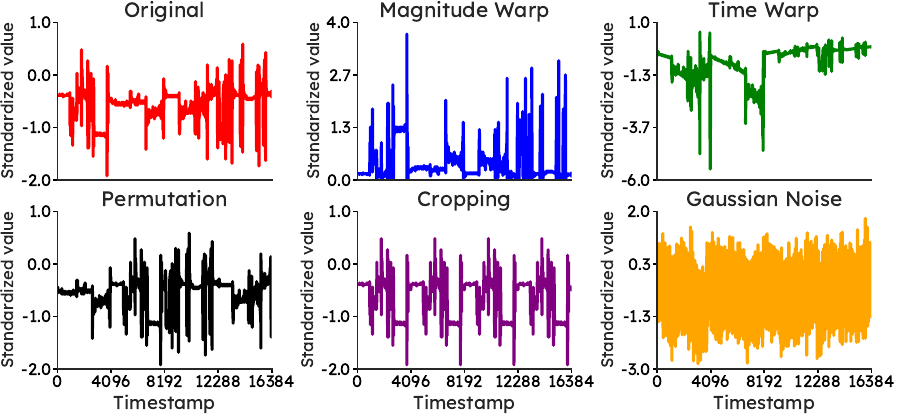}
       \caption{}
       \label{fig:transformations}
    \end{subfigure}
    \caption{\textbf{Surrogate tasks used for self-supervised pre-training}. (\subref{fig:denoising}) \textit{Masked prediction}: grey-shaded areas correspond to zeroed-out time-series portions; the model is tasked with minimizing the distance between the original time-series and the one imputed at the masked areas. (\subref{fig:transformations}) \textit{Transformation prediction}: the figure shows the type of transformations applied to input time-series; given transformed channels, the model was trained to learn which transformation each channel underwent.}
    \label{figure:surrogate_tasks}
\end{figure}

\Cref{figure:surrogate_tasks} illustrates the surrogate tasks we experimented with. In \textbf{masked prediction} (\Cref{fig:denoising}) parts of the input segments are zeroed out by multiplication with a boolean mask sampled as in \citet{zerveas2021transformer} and the model is trained to recover the original input segments. While the model outputs entire segments, only the masked values are taken into account towards the loss computation, that is Root Mean Squared Error (RMSE). The assumption is that the model acquires good representations of the underlying structure of the data when learning to solve this task. Our best model had an error of 0.1347 on the test set (notice that input segments were channel-wise standardized). 

With \textbf{transformation prediction} (\Cref{fig:transformations}) one transformation is sampled from a set and applied to each channel independently and the model learns which transformation each channel underwent, minimizing the channel average categorical cross-entropy (CCE). We used the same transformations as \citet{wu2023transformer}, who experimented with an E4 for a down-stream task of emotion recognition. The rationale is to encourage robustness against signal disturbances introduced with the transformations. The test loss of the selected model was 0.5000.

\subsection*{Target Task Performance Comparison}

We developed two baselines for the target task: (\romannumeral 1) an original E4-tailored end-to-end fully-supervised model (E4mer, \Cref{figure:E4mer}) with an encoder $\mathsf{EN}$, made of convolutional channel embeddings $\mathsf{CEs}$ and a Transformer \cite{vaswani2017attention} representation module $\mathsf{RM}$, followed by an MLP as a classification head $\mathsf{H_{sl}}$; (\romannumeral 2) XGBoost, a competitive CML model, inputting extracted features. For both surrogate tasks introduced above, we took the best performing model and kept the pre-trained $\mathsf{EN}$ while replacing the transform head $\mathsf{H_{ssl}}$ with a new classification head $\mathsf{H_{sl}}$. $\mathsf{EN}$'s parameters were either kept fixed during target-task training, thus recycling $\mathsf{EN}$ as a featurizer and only learning $\mathsf{H_{sl}}$ (\textit{linear readout}), or they were just initialized to the values learned during SSL but then allowed to adapt to the target task along with $\mathsf{H_{sl}}$'s parameters (\textit{fine-tuning}).

\Cref{table:performance} illustrates the performance under each model. While they were all optimized for segment accuracy ($\text{ACC}_{\text{segment}}$, i.e. number of correctly classified segments over total number of segments), since in a clinical scenario a decision needs to be taken at the subject level, we reported subject accuracy ($\text{ACC}_{\text{subject}}$, i.e. number of correctly classified subjects over total number of subjects, where a subject is correctly classified when the majority vote over that subject's segment predictions is in agreement with the subject's true MD state. Note that while accuracy is a suitable metric in our use case as data is perfectly balanced, we also provide complementary metrics (precision, recall, $F_{1}$score, and AUROC) both at the segment and the patient level.

The two baselines both performed to a similar level: while E4mer was superior to XGBoost in terms of $\text{ACC}_{\text{segment}}$ (75.35 vs 72.02), it was trumped by the CML on $\text{ACC}_{\text{subject}}$ (82.81 vs 81.25). Masked prediction pre-training led to a target task performance substantially higher than the baselines, under both metrics. While both linear read-out and tuning dominated over supervised learning, the latter scored the highest performance with an $\text{ACC}_{\text{segment}}$ and an $\text{ACC}_{\text{subject}}$ of 0.8123 and 0.9063 respectively. On the other hand, transformation prediction led to an only modest improvement over E4mer.

\begin{figure}[!t]
    \centering
    \includegraphics[width=0.85\linewidth]{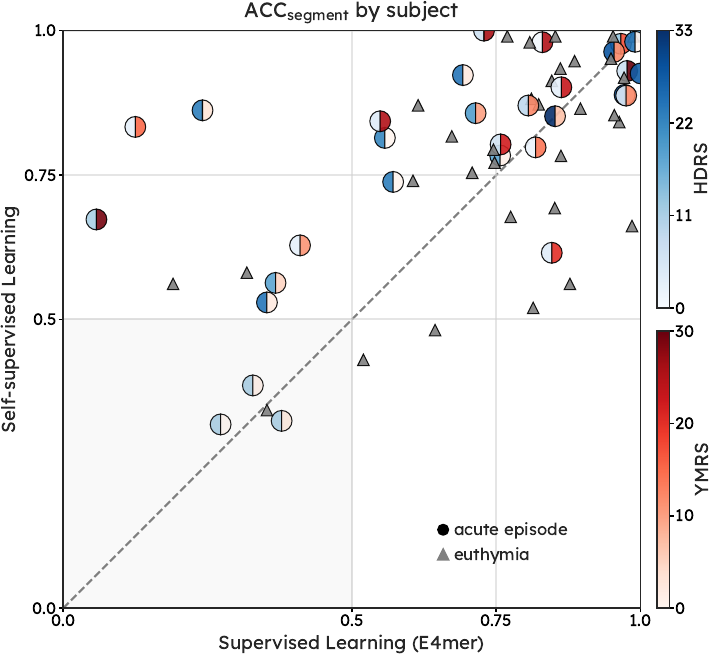}
    \caption{\textbf{Self-supervised learning beats supervised-learning by six more subjects correctly classified}. Segment Accuracy ($\text{ACC}_{\text{segment}}$) under self-supervised learning and supervised learning (E4mer) within each subject's test segments. Subjects in euthymia are represented as triangles while subjects on an acute episode are shown as circles with the left (right) half coloured in blue (red) with a gradient proportional to total sum on the Hamilton Depression Rating Scale-17 (Young Mania Rating Scale), a doctor-administered questionnaire gauging depression (mania) severity. Subjects' position on the x (y) axis corresponds to their proportion of recording segments correctly classified by supervised (self-supervised) learning. Note that a subject's majority vote over their segments is in agreement with the subject's true mood state when the proportion of correctly classified segments from that subject is greater than 0.5. HDRS (YMRS) range showed on the colorbar refers to values scored in the INTREPIBD sample, while the total score in general can range between [0-52] ([0-60]).
    }
    \label{figure:accuracies}

\end{figure}

Comparison of the best SSL with its supervised-learning counter-part in terms of $\text{ACC}_{\text{segment}}$ by subject (\Cref{figure:accuracies}) suggests that there are only two (euthymic) individuals misclassified by SSL but correctly classified by the supervised E4mer. On the other hand, supervised-learning mis-predicts eights individuals that SSL gets right. Patients on an MD acute episode are shown as dots with a color gradient proportional to their total score on the Hamilton Depression Rating Scale-17 \cite{hamilton1960rating} (left half) and Young Mania Rating Scale \cite{young1978rating} (right half), two clinician-administered questionnaires tracking depression and mania severity respectively. Subjects on an acute episode misclassified by supervised learning include patients with a severe depressive (or manic) symptomatology. Notably, both SSL and supervised-learning fail on four subjects, including three patients on an acute episode with relatively moderate severity. 

\begin{table}[!t]
    \caption{\textbf{Masked prediction self-supervised pre-training comfortably outperformed end-to-end self-supervised learning while also dominating over other self-supervised approaches}. Performance in differentiating a mood disorder acute episode from euthymia across different models. Note that data is perfectly balanced in terms of segment classes and subject classes. While this justifies the use of accuracy as a metric, we also herewith report segment and subject level precision, recall, $F_{1}$score, and area under the ROC curve. At the subject level, the predicted class was the result of a majority vote over that subject's segments, while the predicted probabilities under each class were derived by summing segments' predicted probabilities for that subject and normalizing by the corresponding segments number.
    CL: contrastive learning; FT: fine-tuning; LR: linear read-out; MP: masked-prediction; SL: supervised learning; SSL: self-supervised learning; TP: transformation prediction.    
    } \label{table:performance}
    \begin{center}
    \begin{small}
    \begin{sc}
    \setlength{\tabcolsep}{4pt}  %
    \begin{tabular}{p{0.6cm}ccc|cccccccc}
        \toprule
        \multicolumn{2}{c}{\multirow{2}{*}{Model}} & \multicolumn{2}{c}{ACC} & \multicolumn{2}{c}{precision} & \multicolumn{2}{c}{recall} & \multicolumn{2}{c}{$F_{1}$score} & \multicolumn{2}{c}{AUROC}\\
        \multicolumn{2}{c}{} & {\fontsize{6}{10}\selectfont segment} & {\fontsize{6}{10}\selectfont subject} & {\fontsize{6}{10}\selectfont segment} & {\fontsize{6}{10}\selectfont subject} & {\fontsize{6}{10}\selectfont segment} & {\fontsize{6}{10}\selectfont subject} & {\fontsize{6}{10}\selectfont segment} & {\fontsize{6}{10}\selectfont subject} & {\fontsize{6}{10}\selectfont segment} & {\fontsize{6}{10}\selectfont subject} \\
        \midrule
        \multirow{2}{*}{SL} & XGBoost & 72.02 & 82.81 & 71.33 & 83 & 72.11 & 81.1 & 71.72 & 82.03 & 72.44 & 83.17 \\
        & E4mer & 75.35 & 81.25 & 73.46 & 80.55 & 75.34 & 82.14 & 74.39 & 81.33 & 75.68 & 82.22\\
        \midrule
        \multirow{4}{*}{SSL} & MP (LR) & 77.53 & 87.50 & 78.34 & 88.6 & 77.41 & 88 & 77.87& 88.3 & 78.02 & 89.2\\
        & MP (FT) & \textbf{81.23} & \textbf{90.63} & \textbf{80.91} & \textbf{90.11} & \textbf{82.00} & \textbf{92.87} & \textbf{81.45} & \textbf{91.47} & \textbf{82.02} & \textbf{93.11} \\
        & TP (LR) & 71.16 & 81.25 & 72.12& 82.44 & 72.01 & 82.31 & 72.06& 82.37 & 71.89& 84.12 \\
        & TP (FT) & 75.69 & 84.38 & 75.41& 82.11 & 74.79& 83.90 & 75.10&X 83 & 75.21& 84.32 \\
        \bottomrule
    \end{tabular}
    \end{sc}
    \end{small}
    \end{center}
\end{table}
\noindent

\subsection*{Ablation analyses \& learned representations}

Towards elucidating key contributors to the viability of SSL, besides comparing different pretext task designs, we studied how a progressive shrinkage of the available unlabelled data might impact on the performance of our best SSL model. Thus, keeping the best hyperparameters configuration from the experiments above, we re-trained from scratch the SSL model on masked prediction upon resampling the unlabelled data and then fine-tuned it on the target task. Note that resampling was stratified by unlabelled dataset and that the entire target task training set was kept for pre-training under any resampling; this is because pre-training on the training set can be always done at no extra cost in terms of data acquisition. In other words, a resampling percentage of 0\% means that self-supervised pre-training used the target training set only. The following resampling percentages, 80\%, 60\%, 40\%, 20\%, 0\%, resulted in the following values of $\text{ACC}_{\text{segment}}$ ($\text{ACC}_{\text{subject}}$): 81.00, 79.09, 75.16, 74.88, and 74.16 (89.06, 89.06, 85.94, 85.94, and 82.81) respectively. The Pearson correlation coefficient between unlabelled data resampling percentage and $\text{ACC}_{\text{segment}}$  ($\text{ACC}_{\text{subject}}$) is 95.41 (96.14) indicating strong dependence between performance and unlabelled data availability.

\begin{figure}[!t]
    \centering
    \includegraphics[width=1\linewidth]{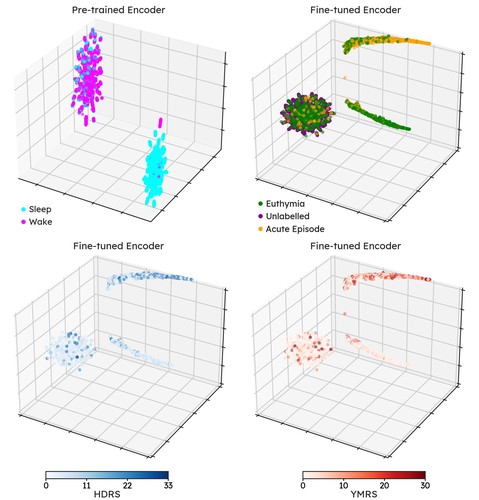}
    \caption{\textbf{Reassuringly, the learned embeddings seem to have captured meaningful semantics about the underlying context}. Top left: embeddings from the encoder pre-trained on mask prediction map sleep and wake segments to different parts of the latent space. Top right: embeddings from the encoder fine-tuned on the target task show that segments from the unlabelled open-access datasets, which presumably do not contain subjects on an acute mood episode, tend to cluster with part of the segments from patients in euthymia. Bottom left (right): embeddings from the fine-tuned encoder show a gradient in symptoms' severity across target task segments, as revealed by Hamilton Depression Rating Scale-17 (Young Mania Rating Scale) total score. Note that unlabelled segments are not showed in the bottom left (right) plot and that the HDRS (YMRS) range showed on the colorbar refers to values scored in the INTREPIBD sample, while the total score in general can range between [0-52] ([0-60])
    }
    \label{figure:representations}
\end{figure}

Lastly, we visualized the representations learned by the encoder $\mathsf{EN}$ part of our best performing models, in order to gain further insights. As the $\mathsf{EN}$'s output had dimensionality ($B$=segments \#, $N$=timestamps \#, $D$=Transformer's model dimension), for visualization purposes we averaged out the $D$ axis and then employed  UMAP \cite{mcinnes1802umap}, a powerful nonlinear dimensionality reduction technique, to embed the resulting $N$-dimensional data points into three dimensions. The top-left plot of \Cref{figure:representations} shows the representations learned during self-supervised pre-training with masked prediction. The segments herewith shown are the target task test segments along with an equal number of segments belonging to the same sessions but taken from sleep state, which the SSL model was never exposed to during training. Wake and sleep segments have different embeddings, suggesting that the model captured this structure in the physiological data: a Gaussian mixture model indeed recovered two clusters, one with predominantly sleep segments (82.66\%) and the other with a majority of wake segments (95.58\%). It should be noted that sleep and wake naturally have quite different semantics with respect to physiological data and the algorithm we employed for sleep/wake differentiation (\textit{Van Hees}~\cite{van2015novel}) uses a simple heuristic defining sleep as a sustained lack of significant changes in the acceleration angle. The top-right and bottom plots of \Cref{figure:representations} illustrates the representations from the SSL model upon fine-tuning on the target task. The top-right scatter plot displays the target task test segments as well as pre-training validation set segments (except for the pre-training segments from the INTREPIBD collection). The latter group of segments we assumed as taken from subjects without an MD acute episode and, arguably, most even without any MD historical diagnosis, since the open-access datasets we found did not select for patients with an MD. The plot shows three clusters whose composition, as recovered with a Gaussian mixture model, is as follows: 1) 79.26\% acute episode, 20.73\% euthymia, 2) 74.16\% euthymia, 25.84\% acute episode, and 3) 91.01\% unlabelled segments, 7.96\% euthymia, and 1.02\% acute episode. The bottom plots in \Cref{figure:representations} show target task segments test segments only (no unlabelled segment), coloured with a gradient proportional to symptoms' severity, as assessed with Hamilton Depression Rating Scale-17 \cite{hamilton1960rating} and Young Mania Rating Scale \cite{young1978rating}. Embeddings would seem to suggest a progression in symptoms' severity across the two clusters of segments on the right of the scatter plot.

\section*{Discussion}

Personal sensing is likely to play a key role in healthcare supply, creating unprecedented opportunities for patient monitoring and just-in-time adaptive interventions \cite{birk2022digital}. Towards delivering on this promise, expert annotation is a major obstacle; this is especially the case with mood disorders (MDs), wherein data annotation is particularly challenging and time-consuming, considering the nature of the disorder. To the best of our knowledge, we are the first to show that self-supervised learning (SSL) is a viable paradigm in personal sensing for MDs, mitigating the annotation bottleneck thanks to the repurposing of existing unlabelled data collected in settings as different as subjects playing Super Mario \cite{svoren2020toadstool}, taking university exams \cite{amin2022wearable_wesd}, or performing physical exercise \cite{gashi2022multidevice_weee}.

We took on a straightforward yet fundamental task, i.e. distinguishing acute episode from euthymia. Timely recognition of an impending mood episode in someone with an historical MD diagnosis, regardless of the episode polarity (depressive, manic, or mixed), may indeed enable pre-emptive interventions and better outcomes \cite{joyce2016treatment}. Our results suggest that, with a sample size on the order of magnitude that is typical of studies into personal-sensing for MD, a modern deep-learning fully-supervised pipeline (E4mer) may offer no substantial improvements over a simpler classical-machine-learning (CML) algorithm (XGBoost), despite higher development and computational costs. On the other hand, the accumulation and repurposing of existing unlabelled datasets for a self-supervised learning (SSL) pre-training phase leads to a confident margin of improvement: $\text{ACC}_{\text{segment}}$ ($\text{ACC}_{\text{subject}}$) improves by 7.8\% (11.54\%) relatively to the fully-supervised E4mer, with 6 (out of 64) more subjects correctly classified.

Our findings further show that careful choice of the pretext task, as well-documented in the literature on SSL \cite{zhang2023self}, is key towards learning useful representations for the downstream target task. Unlike masked prediction, improvement, if any at all, from transformation prediction was in fact only modest. This is not saying that such pretext task may in general fail to deliver on acute episode vs. euthymia differentiation. Indeed, the specific transformations we implemented, borrowed from \citealt{wu2023transformer}, may have been suboptimal for our downstream task, pointing to the importance of domain knowledge (including clinical expertise) in pretext task design. Lastly, while SSL relaxes dependence on large annotated datasets, we note that its success may still be a function of unlabelled dataset size, as ablation analyses showed a strong dependence of performance on pre-training data availability.

We acknowledge the following limitations to this study. We deliberately chose the simplest task that has clinical relevance in personal sensing for MDs since our focus was on SSL; however, we appreciate that a more fine-grained MD description, beyond a simple acute episode vs euthymia binary classification, may add further clinical value. As the literature on SSL is expanding at a fast pace, a thorough search of different approaches was beyond the scope of this work. We acknowledge that other pretext tasks can be deployed and while the architectural choice may have an impact on SSL, we settled for just one reasonable, modern model design with a Transformer \cite{vaswani2017attention} as workhorse for representation learning.

\textbf{Conclusion} - This work shows that self-supervised learning is a promising paradigm for mitigating the annotation bottleneck, one of the major barriers towards the development of AI-powered clinical decision support systems using personal sensing to help monitor mood disorders, thus enabling early interventions. The collection and pre-processing of open-access unlabelled datasets that we curated (E4SelfLearning) can foster future research into self-supervised learning, therefore advancing the translation of personal sensing into the clinical practice.

\textbf{Future directions} - As our findings indicate that the choice of pretext task has a significant impact on target task performance, further efforts should be put into pretext task design. Indeed, while masked prediction is a general purpose strategy inspired by the great success of BERT \cite{devlin2018bert} in Natural Language Processing, literature on self-supervised learning \cite{zhang2023self} suggests that domain knowledge may help tailor the pretext task to the specific use case. A promising approach we have not explored is contrastive learning \cite{kumar2022contrastive}, which indeed relies on domain knowledge into how augmented views of in input are created, especially since most experience today is in Computer Vision and Natural Language Processing while physiological multivariate time-series are relatively unexplored.

\section*{Methods}

\subsection*{Data Pre-processing}

Our pre-processing encompassed the following sequential stages: on-/off-body detection, sleep/wake detection, segmentation, and (when preparing data for CML models) features extraction.

During free-living wear, subjects might remove their device or contact to the wrist might be otherwise suboptimal. As a result, off-body periods can be erroneously mistaken for periods of sleep or sedentary behaviour, due to the shared feature of an absence of movement. Signal discontinuity in biopotentials, such as electrodermal reactivity (EDA), due to a lack of skin contact can be reliably leveraged  to detect non-wear periods. Similarly to \cite{vieluf2021twenty, nasseri2020signal} we considered measurements smaller than 0.05 $\mu$S as indicative of off-body status. Furthermore, as we noticed occurrences of values greater than the EDA sensor range (i.e. 100 $\mu$S \cite{emaptics}), as well as instances of TEMP values outside the physiological range (30-40 \textit{C}$^{\circ}$), we set both to off-body. 

As physiological data vary wildly across sleep and wake, we used sleep-wake detection as a form of data-cleaning to reduce the variance in the signal and considered wake time only in our analyses, especially as most publicly available datasets recorded in wake conditions. We opted for the algorithm by \citealt{van2015novel} (\textit{Van Hees}) which was reported as the best performing in a recent benchmark study on sleep-wake detection (average F1-score 79.1) \citep{patterson202340}. Like most non-proprietary algorithms, \textit{Van Hees} uses triaxial acceleration (ACL) and, specifically, it relies on a simple heuristic defining sleep with the absence of change in arm angle greater than 5 degrees for 5 minutes or more. To accommodate this rule, wherever on-body sampling cycles did not constitute unbroken sequences of at least 5-minute duration, all the measurements in that period were considered as off-body and discarded from further analysis. Median (interquartile range) number of hours marked as off-body, sleep, and wake were respectively 7.73 (13.45), 16.18 (8.43), 18.11 (14.86) in the INTREPIBD cohort, and 0.70 (2.44), 0.00 (0.15), 2.00 (3.28) in the E4SelfLearning collection.

Wake time from each recording was then segmented with a sliding window, whose segment length ($\omega$) and step-size ($\Delta\omega$) (in wall-time seconds) we set to $2^{9}$ and $2^{7}$, respectively. This approach, also referred to window slicing \citep{cui2016multi}, is a common form of data augmentation in time-series classification as multiple segments are produced from a single recording, each one marked with the same label, and is common in personal sensing for MDs. Previous relevant works \cite{cote2022long, pedrelli2020monitoring} defined $\omega$ ($\Delta\omega$) based on clinical intuition and convenience with respect to the available data. Our choice was guided by a similar principle: $2^{9}$, conveniently a power of 2, arguably provides a clinically reasonable observation time, as shorter windows may fail to capture slow-changing dynamics, and maximizes the number of segments available for SSL as some datasets only contain comparatively short recordings.

Recording segments constituted our basic unit of analyses and for the purpose of the target task segments from the same recording all shared the same ground truth label (i.e. either acute episode or euthymia). When fed to deep-learning models, segments were channel-wise standardized by subtracting the mean and dividing by the standard deviation. Such statistics were learned from the target task training set or, in the case of SSL, its aggregation with unlabelled data. Only ACL, BVP, EDA, and TEMP were considered in deep-learning models as HR and IBI are features derived from BVP through a proprietary algorithm. On the other hand, when using CML, handcrafted features were extracted from segments using \textit{FLIRT} \citep{foll2021flirt}, a popular open-access feature extraction toolkit for Emapatica E4. Note that a single row of features per segment was extracted, in other words, the window size parameter in \textit{FLIRT} was set equal to $\omega$. We used all features available through this package and any missing value was handled with mean imputation.

\subsection*{Data splits \& Metrics}

In SSL experiments, we split data with a ratio of 85/15 into train and validation set, partitioning recordings across the two sets. As for the target task,  we investigated a \textit{time-split} scenario therefore splitting each recording into train/val/test again with a ratio of 70/15/15 along recording time, thus testing generalization across future time points. We made sure that segments with overlapping motifs at the border between target task splits (resulting from using a sliding window with $\Delta\omega < \omega$) were confined to one split only, thus ultimately producing segments\# 18896/3904/4128 for train/val/test. The target task validation test doubled as a test set for SSL pretext task generalization performance estimate. \textit{Time-split} scenario is common in personal sensing for MDs (e.g. \cite{pedrelli2020monitoring, corponi2023automated}) and indeed, despite efforts towards learning subject-invariant representations \cite{ozdenizci2020learning, cheng2020subject}, cross-subjects generalization remains an unsolved challenge so that personal sensing systems typically require access to each subject's physiological data distribution at training time \cite{sabry2022machine}.

The target task is time-series binary classification. As expected in free-living wear, total wear time as well as on-body and wake time vary across subjects (and, as a result, so does the number of segments). We extracted an equal number of segments from each class (acute episode and euthymia) and optimized models on the target task for segment level accuracy ($\text{ACC}_{\text{segment}}$), i.e. $\frac{1}{N} \sum_{i=1}^{N} \mathbbm{1}(\hat{y}_{i} = y_{i})$, where ${i}$ indexes the segments. Secondarily, in order to provide a subject level perspective, we reported the subject accuracy ($\text{ACC}_{\text{subject}}$), i.e. $\frac{1}{S} \sum_{s=1}^{S} \mathbbm{1}(\hat{y}_{s} = y_{s})$, where $y_{s}$ is the ground truth mood state of the $s^{th}$ subject, which is constant across all $s^{th}$ subject's recording segments, and $\hat{y}_{s}$ is a majority vote on the $s^{th}$ subject, corresponding to the majority predicted class across the $s^{th}$ subject's recording segments.

\subsection*{Machine learning models}

We developed two baselines for the target task: 1) an E4-tailored deep-learning pipeline inputting raw recording segments (E4mer) and 2) a CML model using handcrafted features extracted with FLIRT from recording segments. We then assessed what boost in performance, if any, a self-supervised pre-training phase might deliver, where the SSL models share the same building blocks as the E4mer. 

\subsubsection*{Baseline Models}

\textbf{E4mer} - This is an artificial neural network discriminative classifier modelling the probability of an MD acute episode given a recording segment, $P(y_{i}=1 \mid x_{i})$. As shown in \Cref{figure:E4mer}, our E4mer has three sequential blocks: 1) channel embeddings ($\mathsf{CE}$) set in parallel, consisting of 1D same Convolutions with a kernel size equal to the channel sampling frequency, followed by Gelu activation, 1D BatchNorm, and 1D MaxPooling using the channel sampling frequency as both kernel size and step size so that each channel embedding output had the same dimensionality and could be conveniently concatenated with the others before being passed onto 2) a Transformer \cite{vaswani2017attention} representation module ($\mathsf{RM}$), and 3) a Multi-Layer Perceptron (MLP) classification head ($\mathsf{H_{sl}}$). The $\mathsf{CE}$ extract features from the input E4 channels and are designed to handle channels sampled at different frequencies, the $\mathsf{RM}$, powered by multi-head self-attention, learns contextual representations of the input tokens (time-stamps in our case) within a recording segment, the $\mathsf{H_{sl}}$, lastly, maps such representations onto a label space appropriate for a binary classification. The E4mer is trained to minimize the binary cross-entropy (BCE) loss between acute episode/euthymia predictions and the corresponding ground truth.

\textbf{XGBoost} - This is a state-of-the-art CML algorithm, showing good performance in previous studies into personal sensing \cite{tazawa2019actigraphy, jacobson2019digital}. XGBoost operates by building an ensemble of decision trees sequentially, where each tree is designed to correct the errors of the previous ones. Concretely, with each iteration, a new tree is fit to the pseudo-residuals from the previous tree(s), i.e. the negative gradient of a differentiable loss function (logistic loss in our case) evaluated at the previous model's predictions.

\subsubsection*{Self-supervised learning schemes}

SSL schemes rely on devising a pre-text task, for which a (relatively) large amount of unlabelled data is available, conducive to learning, during a pre-training phase, representations useful to solve the down-stream target task \citep{ericsson2022self}. What defines an SSL paradigm is thus its pretext task, consisting of a process, $\mathit{P}$, to generate pseudo-labels and an objective to guide the pre-training. An SSL model typically consists of (\romannumeral 1) an encoder $\mathsf{EN}\left ( x; \theta \right ):\mathit{X} \to \mathit{V}$, learning a mapping from inputs views $x \in X$ to a representation vector $v \in R^{d}$ and (\romannumeral 2) a transform head $\mathsf{H_{ssl}}\left ( v; \xi \right ):\mathit{V} \to \mathit{Z}$ projecting the feature embedding into a label space $z \in R^{d\prime}$ compatible with the pretext task at hand. When solving the target task, the pre-trained encoder $\mathsf{EN}$ is retained as a partial solution to the target problem, whereas the pre-trained transform head $\mathsf{H_{ssl}}$ is discarded and replaced with a new one $\mathsf{H_{sl}}$. Then, $\mathsf{EN}$'s parameters $\theta$ may be kept fixed and only $\mathsf{H_{sl}}$'s parameters may be learned on the target task. This approach, often referred to as \textit{linear readout} amounts to treating $\mathsf{EN}$ as a frozen feature extractor. Alternatively, instead of just training a new head, the entire network $\mathsf{EN}\left ( \mathsf{H_{sl}}\left ( \cdot \right ) \right )$ may be retrained on the target task, initializing $\mathsf{EN}$'s parameters $\theta$ to the values learned during self-supervised pre-training, a paradigm known as \textit{fine-tuning}. Our SSL models used the same architecture as the E4mer, that is an encoder $\mathsf{EN}$, consisting of convolutional $\mathsf{CEs}$ followed by a Transformer $\mathsf{RM}$, and an MLP for the transform head $\mathsf{H_{ssl}}$.

The success of SSL methods largely comes from designing appropriate pretext tasks that will produce representations useful for the down-stream target task. This usually involve domain-knowledge into the target task. We herewith investigated how different pretext tasks affected downstream performance, experimenting with two popular SSL routines that showed success in other applications: masked prediction and transformation prediction.

\textbf{Masked predictions} This family of SSL methods is characterized by training the model to impute data removed or corrupted by $\mathit{P}$. It relies on the assumption that context can be used to infer some types of missing information in the data if the domain is well modelled. This strategy was popularized by the huge success of BERT \citep{devlin2018bert} in NLP applications, and a first adaptation to multi-variate time-series classification was proposed by \citealt{zerveas2021transformer}. Similarly to their implementation, for each segment channel we sampled a boolean mask whose 0s and 1s sequence followed a geometric distribution with mean $l_{m}$ and $l_{u} = \frac{1-r}{r} l_{m}$ respectively, where $r$ is the masking ratio. As in \cite{zerveas2021transformer}, the average length (in wall-time seconds) of the 0s sequences ($l_{m}$) and the proportion of masked values ($r$) were set to 3 and 0.15 respectively. Each segment channel was then multiplied by its corresponding mask, effectively setting to 0 some of the channel recorded measurements, and inputted to a model which was tasked to recover the original channel values. This was done minimizing the Root Mean Squared Error (RMSE) between the masked original value $x(t, c)$ and its reconstruction outputted by the network $\hat{x}(t, c)$: 

\begin{linenomath*}
\begin{align*}
\mathcal{L}_{\mathsf{RMSE}}=\sqrt{\frac{1}{|M|} \sum_{t \in M} \sum_{c \in M}(\hat{x}(t, c)-x(t, c))^2}
\end{align*}
\end{linenomath*}

where $c$ and $t$ respectively index the channels and the timestamps of the 0s values in the masks $M$ and $|M|$ is the total number of 0s sampled, i.e. the masks' cardinality.

\textbf{Transformation prediction} We followed the implementation by \citealt{wu2023transformer} which used SSL for a target task of emotion recognition with E4 recordings. In brief, for each channel one of six transformations (i.e. identity, Gaussian noise addition, magnitude-warping, permutation, time-warping, and cropping) was sampled uniformly at random and then applied. The transformed segment was then inputted to a model, which was tasked to guess for each channel which one of the six transformations was applied. This amounted to a multitask multi-class classification where the model was trained to minimize channel average categorical cross-entropy (CCE):

\begin{linenomath*}
\begin{align*}
\mathcal{L}_{\mathsf{CCE}_{Multitask}}=\frac{1}{C} \sum_{c=1}^{C} \sum_{j=1}^{T} -\mathbbm{1}_{c,j} \cdot \log(p_{c,j})
\end{align*}
\end{linenomath*}

where $c$ indexes the channels, $j$ the transformations, $\mathbbm{1}_{i,j}$ is an indicator taking value 1 when $j$ is correct transformation for the channel $c$, 0 otherwise, and $p_{c,j}$ denotes the predicted probability that transformation $j$ was applied to channel $c$. By solving this task, the authors in \cite{wu2023transformer} argue that the model learns representations robust to disturbances in the magnitude and time domains.

\subsection*{Tuning}

Hyperparameters search for all models was carried out with Hyperband Bayesian optimization \cite{li2017hyperband}. For the target task, we selected the setting yielding the highest $\text{ACC}_{\text{segment}}$ in the validation set, whereas in self-supervised pre-training we selected hyperparameters associated with the lowest relevant loss in the validation pre-training set. \Cref{HPs} shows the hyperparameters search space and the best configuration across all models. Deep-learning models were trained with AdamW optimizer for a maximum of 300 epochs, with a batch size of 256. Moreover, to speed up the training and search procedure, we employed an early stopping learning rate scheduler: we reduce the learning rate $\alpha_\textsc{lr} = 0.3\alpha_\textsc{lr}$ if the model has not improved in its validation performance after 10 consecutive epochs; we terminate the training procedure if the model has not improved after 2 learning rate reductions. Dropout \cite{srivastava2014dropout} and weight decay were added to prevent overfitting.

\section*{Code availability}

The codebase developed for this work is available at \href{https://github.com/april-tools}{link to be released upon acceptance for publication}. Python 3.10 programming language was used where deep-learning models and XGBoost were implemented in PyTorch \cite{NEURIPS2019_9015} and XGBoost \cite{ChenXGBoost} respectively, while hyperparameter tuning was performed in both cases with \href{https://wandb.ai/}{Weight \& Biases}~\cite{wandb}. All deep-learning models were trained on a single Nvidia A100 GPU.

\section*{Data availability}

The E4SelfLearning collection is available at \href{https://github.com/april-tools}{link to be released upon acceptance for publication}. Data in de-identified form from the INTREPIBD study may be made available from the corresponding author upon reasonable request.

\section*{Ethics and confidentiality}\label{methods:ethics}

The INTREPIBD study was conducted in accordance with the ethical principles of the Declaration of Helsinki and Good Clinical Practice and the Hospital Clinic Ethics and Research Board (HCB/2021/104). All participants provided written informed consent prior to their inclusion in the study. All data were collected anonymously and stored encrypted in servers complying with all GDPR and HIPAA regulations. As regards other studies included in the present work, we refer to the relevant publications.

\section*{Acknowledgments}

We acknowledge the contribution of all the participants of the study.

F.C. and B.M.L. are supported by the United Kingdom Research and Innovation (grant EP/S02431X/1), UKRI Centre for Doctoral Training in Biomedical AI at the University of Edinburgh, School of Informatics. For the purpose of open access, the author has applied a creative commons attribution (CC BY) licence to any author accepted manuscript version arising.

G.A. is supported by a Rio Hortega 2021 grant (CM21/00017) and M-AES mobility fellowship (MV22/00058), from the Spanish Ministry of Health financed by the Instituto de Salud Carlos III (ISCIII) and co-financed by the Fondo Social Europeo Plus (FSE+).

I.G. thanks the support of the Spanish Ministry of Science and Innovation (MCIN) (PI23/00822) integrated into the Plan Nacional de I+D+I and cofinanced by the ISCIII-Subdirección General de Evaluación y confinanciado por la Unión Europea (FEDER, FSE, Next Generation EU/Plan de Recuperación Transformación y Resiliencia PRTR); the Instituto de Salud Carlos III; the CIBER of Mental Health (CIBERSAM);  and the Secretaria d’Universitats i Recerca del Departament d’Economia i Coneixement (2021 SGR 01358), CERCA Programme / Generalitat de Catalunya as well as the Fundació Clínic per la Recerca Biomèdica (Pons Bartran 2022-FRCB PB1 2022). 

A.H.Y.'s independent research is funded by the National Institute for Health and Care Research (NIHR) Maudsley Biomedical Research Centre at South London and Maudsley NHS Foundation Trust and King's College London. The views expressed are those of the author(s) and not necessarily those of the NIHR or the Department of Health and Social Care. For the purposes of open access, the author has applied a Creative Commons Attribution (CC BY) licence to any Accepted Author Manuscript version arising from this submission.

D.H.M. is supported by a Juan Rodés JR18/00021 granted by the Instituto de Salud Carlos III (ISCIII).

A.V. is supported by the "UNREAL" project (EP/Y023838/1) selected by the ERC and funded by UKRI EPSRC.

\section*{Authors contributions} 

F.C. conceived of the study, proposed the methodology, developed the software codebase for the analyses, prepared the manuscript, and curated data collection. B.M.L. contributed to codebase development and manuscript writing.  G.A., C.V.P., A.M., I.P., M.V., I.G.F., A.B., and M.G. collected the data for the INTREPIBD study. E.V., A.H.Y., S.L., and H.W. critically reviewed the manuscript and provided feedback on the clinical side. D.H.M. is the co-ordinator of the INTREPIBD study and critically reviewed the manuscript. A.V. supervised this study and contributed to the study design, methodology development, and manuscript writing.

\section*{Competing interests} 

G.A. has received CME-related honoraria, or consulting fees from Janssen-Cilag, Lundbeck, Lundbeck/Otsuka, and Angelini, with no financial or other relationship relevant to the subject of this article. 

I.G. has received grants and served as consultant, advisor or CME speaker for the following identities: ADAMED, Angelini, Casen Recordati, Esteve, Ferrer, Gedeon Richter, Janssen Cilag, Lundbeck, Lundbeck-Otsuka, Luye, SEI Healthcare, Viatris outside the submitted work. She also receives royalties from Oxford University Press, Elsevier, Editorial Médica Panamericana.

All authors report no financial or other relationship relevant to the subject of this article.

\bibliographystyle{apalike}

\clearpage

\appendix

\section{Supplementary information} \label{appendix}

\setcounter{figure}{0}    
\setcounter{table}{0}

\subsection{Hyperparameter search space and final configuration.} \label{HPs}

\begin{table}[ht]
    \caption{\textbf{Masked Prediction}} \label{table:hp_masked_prediction}
    \begin{center}
    \begin{small}
    \begin{sc}
    \begin{tabular}{llr}
        \toprule
        hyperparameter & search space & final value \\
        \midrule
        learning rate $\alpha_\text{lr}$ & uniform, min: 0.0001, max: 0.01 & 0.009 \\
        weight decay & uniform, min: 0, max: 1 & 0.0572 \\
        \midrule
        Channel embeddings & & \\
        num. filters & categorical, $2^{i}$ for $i\in \left\{2,3,4\right\}$ & $2^{4}$ \\
        \midrule
        Representation module & & \\
        d model & categorical, $2^{i}$ for $i\in \left\{5,6,7,8,9\right\}$ & $2^{8}$ \\
        num. heads & uniform, min: 1, max 4 & 2 \\
        num. blocks & uniform, min: 1, max: 4 & 4 \\
        attention dropout & uniform, min: 0, max: 0.5 & 0.2699 \\
        drop path & uniform, min: 0, max: 0.5 & 0.0034 \\
        mlp dim. & uniform, min: 32, max: 256, interval: 8 & 72\\
        mlp dropout & uniform, min: 0, max: 0.5 & 0.0824 \\
        disable bias & bernoulli & 1 \\
        \bottomrule
    \end{tabular}
    \end{sc}
    \end{small}
    \end{center}
\end{table}

\begin{table}[ht]
    \caption{\textbf{Transformation Prediction}} \label{table:hp_transformation_prediction}
    \begin{center}
    \begin{small}
    \begin{sc}
    \begin{tabular}{llr}
        \toprule
        hyperparameter & search space & final value \\
        \midrule
        learning rate $\alpha_\text{lr}$ & uniform, min: 0.0001, max: 0.01 & 0.009 \\
        weight decay & uniform, min: 0, max: 1 & 0.101 \\
        \midrule
        Channel embeddings & & \\
        num. filters & categorical, $2^{i}$ for $i\in \left\{2,3,4\right\}$ & $2^{4}$ \\
        \midrule
        Representation module & & \\
        d model & categorical, $2^{i}$ for $i\in \left\{5,6,7,8,9\right\}$ & $2^{9}$ \\
        num. heads & uniform, min: 1, max 4 & 2 \\
        num. blocks & uniform, min: 1, max: 4 & 1 \\
        attention dropout & uniform, min: 0, max: 0.5 & 0.3883 \\
        drop path & uniform, min: 0, max: 0.5 & 0.1783 \\
        mlp dim. & uniform, min: 32, max: 256, interval: 8 & 176 \\
        mlp dropout & uniform, min: 0, max: 0.5 & 0.0310 \\
        disable bias & bernoulli & 1 \\
        \bottomrule
    \end{tabular}
    \end{sc}
    \end{small}
    \end{center}
\end{table}

\begin{table}[ht]
    \caption{\textbf{Masked Prediction - Linear Readout}} \label{table:hp_masked_prediction_readout}
    \begin{center}
    \begin{small}
    \begin{sc}
    \begin{tabular}{llr}
        \toprule
        hyperparameter & search space & final value \\
        \midrule
        learning rate $\alpha_\text{lr}$ & uniform, min: 0.0001, max: 0.01 & 0.0058 \\
        weight decay & uniform, min: 0, max: 1 & 0.8189 \\
        \bottomrule
    \end{tabular}
    \end{sc}
    \end{small}
    \end{center}
\end{table}

\begin{table}[ht]
    \caption{\textbf{Transformation Prediction - Linear Readout}} \label{table:hp_transformation_prediction_readout}
    \begin{center}
    \begin{small}
    \begin{sc}
    \begin{tabular}{llr}
        \toprule
        hyperparameter & search space & final value \\
        \midrule
        learning rate $\alpha_\text{lr}$ & uniform, min: 0.0001, max: 0.01 & 0.0058 \\
        weight decay & uniform, min: 0, max: 1 & 0.7952 \\
        \bottomrule
    \end{tabular}
    \end{sc}
    \end{small}
    \end{center}
\end{table}

\begin{table}[ht]
    \caption{\textbf{Masked Prediction - Fine-tuning}} \label{table:hp_masked_prediction_ft}
    \begin{center}
    \begin{small}
    \begin{sc}
    \begin{tabular}{llr}
        \toprule
        hyperparameter & search space & final value \\
        \midrule
        learning rate $\alpha_\text{lr}$ & uniform, min: 0.0001, max: 0.01 & 0.0010 \\
        weight decay & uniform, min: 0, max: 1 & 0.0232 \\
        \midrule
        Representation module & & \\
        attention dropout & uniform, min: 0, max: 0.5 & 0.4732 \\
        drop path & uniform, min: 0, max: 0.5 & 0.01032 \\
        mlp dropout & uniform, min: 0, max: 0.5 & 0.1209 \\
        \bottomrule
    \end{tabular}
    \end{sc}
    \end{small}
    \end{center}
\end{table}

\begin{table}[ht]
    \caption{\textbf{Transformation Prediction - Fine-tuning}} \label{table:hp_transformation_prediction_ft}
    \begin{center}
    \begin{small}
    \begin{sc}
    \begin{tabular}{llr}
        \toprule
        hyperparameter & search space & final value \\
        \midrule
        learning rate $\alpha_\text{lr}$ & uniform, min: 0.0001, max: 0.01 & 0.0011 \\
        weight decay & uniform, min: 0, max: 1 & 0.7052 \\
        \midrule
        Representation module & & \\
        attention dropout & uniform, min: 0, max: 0.5 & 0.0754 \\
        drop path & uniform, min: 0, max: 0.5 & 0.2065 \\
        mlp dropout & uniform, min: 0, max: 0.5 & 0.2782 \\
        \bottomrule
    \end{tabular}
    \end{sc}
    \end{small}
    \end{center}
\end{table}

\begin{table}[ht]
    \caption{\textbf{E4mer - Supervised Learning}} \label{table:hp_ann_sl}
    \begin{center}
    \begin{small}
    \begin{sc}
    \begin{tabular}{llr}
        \toprule
        hyperparameter & search space & final value \\
        \midrule
        learning rate $\alpha_\text{lr}$ & uniform, min: 0.0001, max: 0.01 & 0.00516 \\
        weight decay & uniform, min: 0, max: 1 & 0.0016 \\
        \midrule
        Channel embeddings & & \\
        num. filters & categorical, $2^{i}$ for $i\in \left\{2,3,4\right\}$ & $2^{2}$ \\
        \midrule
        Representation module & & \\
        d model & categorical, $2^{i}$ for $i\in \left\{5,6,7,8,9\right\}$ & $2^{5}$ \\
        num. heads & uniform, min: 1, max 4 & 2 \\
        num. blocks & uniform, min: 1, max: 4 & 4 \\
        attention dropout & uniform, min: 0, max: 0.5 & 0.1702 \\
        drop path & uniform, min: 0, max: 0.5 & 0.4676 \\
        mlp dim. & uniform, min: 32, max: 256, interval: 8 & 120\\
        mlp dropout & uniform, min: 0, max: 0.5 & 0.1037 \\
        disable bias & bernoulli & 0 \\
        \bottomrule
    \end{tabular}
    \end{sc}
    \end{small}
    \end{center}
\end{table}

\begin{table}[ht]
    \caption{\textbf{XGBoost - Supervised Learning}} \label{table:hp_xgboost_sl}
    \begin{center}
    \begin{small}
    \begin{sc}
    \begin{tabular}{llr}
        \toprule
        hyperparameter & search space & final value \\
        \midrule
        colsample by tree & uniform, min: 0.1, max: 1 & 0.9639 \\
        gamma & uniform, min: 0, max: 10 & 0.7854 \\
        learning rate & uniform, min: 0.001, max: 0.3 & 0.2848 \\
        max depth & uniform, min: 3, max: 10, interval: 1 & 9 \\
        min child weight & uniform, min: 0.01, max: 10 & 8.7346 \\
        n estimators & uniform, min: 5, max: 50, interval: 1 & 50 \\
        reg alpha & uniform, min: 0, max: 10 & 0.5811 \\
        reg lambda & uniform, min: 0, max: 10 & 4.2859 \\
        subsample & uniform, min: 0.1, max: 1 & 0.9424 \\
        \bottomrule
    \end{tabular}
    \end{sc}
    \end{small}
    \end{center}
\end{table}

\end{document}